\documentclass[11pt]{article}

\usepackage[final]{acl}

\usepackage{times}
\usepackage{latexsym}
\usepackage[T1]{fontenc}
\usepackage[utf8]{inputenc}
\usepackage{microtype}
\usepackage{inconsolata}
\usepackage{graphicx}
\usepackage{multirow}
\usepackage{xcolor}
\usepackage{booktabs}

\usepackage{tcolorbox}
\usepackage{array}
\newcolumntype{L}[1]{>{\raggedright\arraybackslash}p{#1}}

\newtcolorbox{promptbox}[1][]{
    colback=gray!5,
    colframe=gray!100,
    fonttitle=\bfseries\small,
    title=#1,
    boxrule=0.5pt,
    left=4pt, right=4pt, top=4pt, bottom=4pt,
    fontupper=\small\ttfamily\raggedright
}

\title{\textsc{PatientAct}: Theory-Grounded Mental Health Client Simulation}

\author{Sahand Sabour\textsuperscript{\rm 1} \quad TszYam NG\textsuperscript{\rm 1} \quad Yaqian Chen\textsuperscript{\rm 2}\quad Guanqun Bi\textsuperscript{\rm 1}\quad Jialu Zhao\textsuperscript{\rm 3}\quad Minlie Huang\textsuperscript{\rm 1}
\\ \\
\textsuperscript{\rm 1} The CoAI Group, DCST, Institute for Artificial Intelligence, Tsinghua University, Beijing, China \\
\textsuperscript{\rm 2}Department of Psychology, Beijing Normal University, Beijing, China \\
 \textsuperscript{\rm 3}Counseling and Psychological Development Guidance Center, Tsinghua University, Beijing, China \\
  \texttt{sahandfer@gmail.com}, \texttt{aihuang@tsinghua.edu.cn}}

\begin{document}
\maketitle
\begin{abstract}
LLM-based simulated clients are increasingly used to train novice counselors, evaluate LLM therapists, and generate synthetic data. 
However, current simulators produce overly cooperative clients that disclose too readily, accept therapeutic reframes without resistance, and resolve core issues within a single session. 
We trace these issues to profiles that lack causal depth and behavioral mechanisms that treat all content as equally accessible. 
We present \textsc{PatientAct}, a framework for client simulation grounded in established clinical theories. 
Our profiles integrate the 5Ps clinical case formulation, providing causal depth without tying the design to any single therapeutic modality.
During simulation, profiles include a dynamic memory layer in which items carry trust thresholds (e.g., symptoms are available early, whereas formative memories require a sustained therapeutic alliance).
At each turn, we model the client's emotional reaction and behavior before generating a response.
If the therapist approaches gated content, \textsc{PatientAct} expresses resistance in terms of quantity, content, and style rather than defaulting to cooperation or a single resistance pattern.
We evaluate our framework on 40 clinical situations and demonstrate that it generates diverse profiles with high clinical plausibility.
Moreover, \textsc{PatientAct} significantly outperforms the baselines, yielding substantial gains in resistance quality and behavioral realism.
Our code and data are publicly available via \url{github.com/Sahandfer/PatientHub}.
\end{abstract}

\section{Introduction}
\label{sec:introduction}
Mental health conditions affect over one billion people worldwide, yet most receive no adequate care \citep{who2025}. 
Bridging this gap requires progress on multiple fronts: training more clinicians, developing AI-assisted therapeutic tools, and building research infrastructure for computational mental health. 
Simulated clients powered by large language models (LLMs) have emerged as a foundational component across all three: they provide scalable practice environments for training novice counselors \cite{wang2024patient, lin2026candormd}, serve as standardized test cases for evaluating LLM therapists \cite{sabour2023chatbot}, and enable the generation of diverse synthetic therapy data for psychology research \cite{liu2025enhanced, li2026synthetic}.
At the core of these applications lies a common requirement: the simulated client must behave realistically enough for the interaction to capture the dynamics of real therapy.

In real therapy sessions, clients do not simply answer questions or disclose information.
They react emotionally to what the therapist says (e.g., with relief, shame, or defensiveness) and behave in ways shaped by a lifetime of learned patterns: deflecting when a topic feels threatening, going quiet when overwhelmed, or cautiously opening up when they feel understood. 
In addition, they selectively share information, revealing surface complaints early while guarding painful memories until trust is established.
And when they resist, they do so in varied ways, each reflecting a different underlying pattern; for instance, going silent, changing the subject, or pushing back directly.
These dynamics distinguish a genuine therapeutic interaction from a cooperative question-answering exchange.

Current LLM-based simulators fall short of capturing these dynamics.
A persistent finding across prior work is that simulated clients are overly cooperative: they disclose too readily, accept therapeutic reframes without resistance, and resolve core issues within a single session \cite{yang-etal-2025-consistent, kim-etal-2025-share-story}.
This pattern undermines downstream applications: therapist trainees do not encounter realistic resistance, LLM-therapist evaluations are inflated by compliant clients, and synthetic data lacks the friction that characterizes real therapeutic dialogue. 
We attribute this problem to several shortcomings in existing simulator designs.

First, prior profile designs lack causal depth.
Existing profiles describe \textit{what} the client thinks and feels by including attributes such as symptoms, beliefs, and coping strategies \cite{wang2024patient, wang2024towards, lee2025psyche}.
However, they do not explain \textit{why} the client has formed such thoughts and feelings: what made this person vulnerable, what triggered the current episode, or what cycles sustain it.
Hence, the resulting simulator can state ``I feel worthless'' but cannot explain how this belief was formed or what in their daily life reinforces it.
Second, prior work either provides the full profile in the system prompt \cite{wang2024patient, lee2025psyche}, thereby making all information immediately accessible, or applies a single control (e.g., a static label such as \textit{Resistance: High}) uniformly across all content \cite{kim-etal-2025-share-story}.
However, in practice, a client may freely describe sleep problems while actively avoiding childhood memories, and this boundary shifts as trust develops within the conversation \citep{srivastava2025trust}.

To address these shortcomings, we present \textsc{PatientAct}, a theory-grounded framework for client simulation. 
Our profiles follow the 5Ps clinical case formulation \citep{johnstone2013formulation}, which traces how a client's vulnerabilities developed, what triggered the current episode, and what maintains the problem.
We complement this formulation with a cognitive layer \citep{beck2020cognitive} that captures in-session thoughts and behaviors, and an interpersonal relational layer \citep{luborsky1998understanding} that shapes how clients relate to others, particularly the therapist.
Unlike prior work, which mainly focuses on a single therapeutic modality, such as cognitive behavioral therapy (CBT) or motivational interviewing (MI), our schema is not tied to any single therapeutic modality (i.e., modality-agnostic).
Prior to the simulation, we divide the profile into a static layer, which is always included in the system prompt, and a dynamic memory layer.
Each memory item is assigned a disclosure threshold, grounded in prior research on therapeutic trust \citep{srivastava2025trust}: surface symptoms are available early, whereas formative memories require sustained therapeutic alliance.
During the simulation, \textsc{PatientAct} determines how the client should emotionally react to the therapist's utterance, and selects a behavior consistent with that reaction and current trust level before generating a 
response.
When the therapist inquires about content that the client is not yet ready to share, \textsc{PatientAct} displays resistance based on \citet{otani1989client}'s taxonomy spanning response quantity (e.g., going silent), content (e.g., changing the subject), and style (e.g., direct pushback).
As a result, the same profile can produce meaningfully different sessions depending on the therapist's behavior.
Our contributions are as follows:
\begin{enumerate}
    \item A \textbf{modality-agnostic profile schema} that integrates clinical case formulation, providing the causal depth missing from prior work.
    \item A \textbf{theory-grounded simulation framework} that processes each therapist's utterance through reaction, behaviors, and a dynamic retrieval pipeline. The client's evolving trust gates disclosure of the profile's content, and resistance spans multiple clinical dimensions rather than a single label.
    \item \textbf{Comprehensive evaluation} demonstrating that our framework produces more realistic client simulations than existing baselines across expert and LLM-based assessments.
\end{enumerate}

\begin{figure*}[t]
    \centering
    \includegraphics[width=\linewidth]{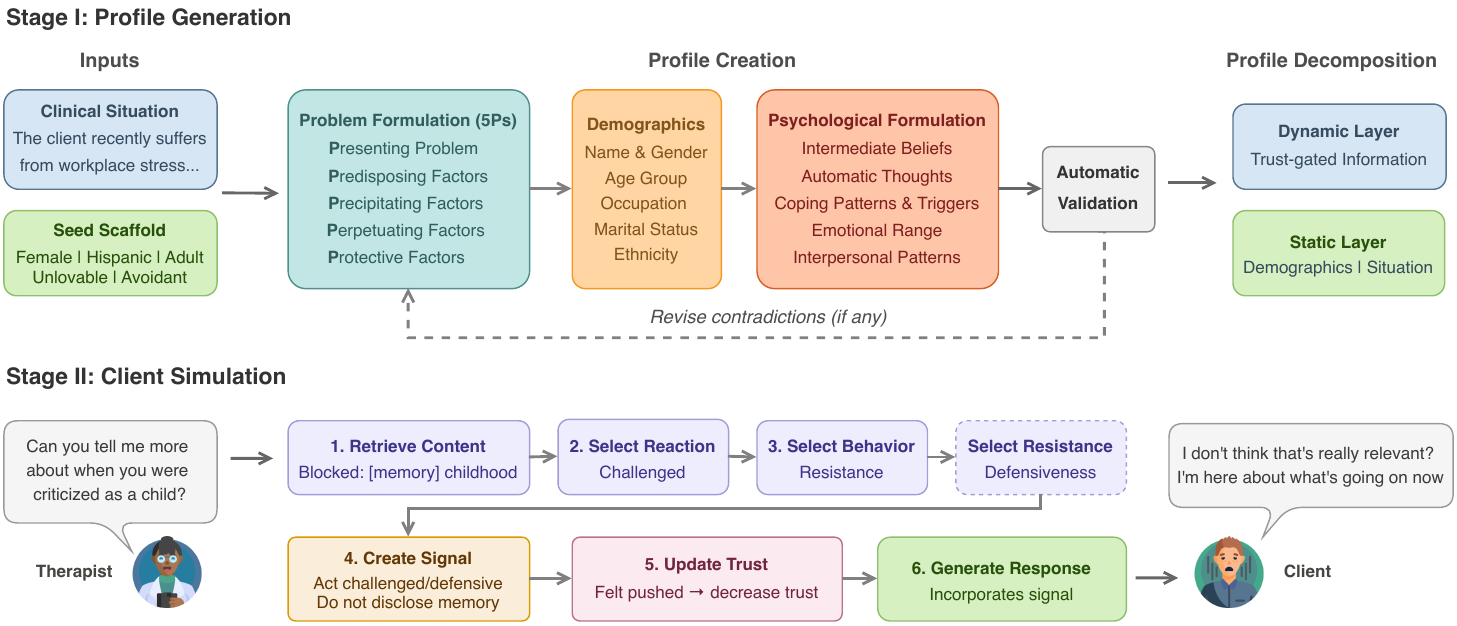}
    \caption{Overview of Our Framework (\textsc{PatientAct}).}
    \label{fig:framework}
\end{figure*}

\section{Related Work}
\label{sec:related-work}
LLM-based patient simulation has emerged as a flexible alternative to scripted standardized patients, leveraging modern LLMs' role-playing and instruction-following capabilities.
Existing approaches differ primarily along two axes: \textit{what} is encoded about the client (profile design) and \textit{how} the desired behavior is enforced during conversation (simulation design). 

\subsection{Profile Design}
A central challenge in patient simulation is determining the information required to model clients and how to structure it.
Early approaches relied on scenario descriptions and simple persona descriptions covering demographics and presenting complaints without a systematic structure \citep{chen2023llm, wang2024towards, 
louie2024roleplay}, causing the LLM to improvise information (e.g., beliefs) during conversations.
A shift toward theory-grounded design began with Patient-$\psi$ \citep{wang2024patient}, which structured profiles around the Cognitive Conceptualization Diagram \citep[CCD;][]{beck2020cognitive} and captured the client's cognitive layer by encoding core beliefs, intermediate beliefs, and coping strategies. 
However, the CCD does not explain how those beliefs were formed or what sustains them.
Subsequent work broadened the scope by including psychiatric history, longitudinal symptom data, and life events \citep{lee2025psyche, liu2025eeyore, wang-etal-2025-annaagent, li2026synthetic}, adding biographical detail but not the causal links between those events and the presenting problem.
Hence, existing profiles encode \textit{what} the client thinks and feels but do not provide sufficient reasons for \textit{why} the client feels this way. 
For instance, they do not account for what made the person vulnerable, what triggered the current episode, or what perpetuates the problem. 
Without this structure, a simulated client cannot coherently explain their history, exhibit maintenance behaviors that a therapist would recognize and work with, or produce realistic resistance: all profile content carries equal weight, with no basis for distinguishing what the client would share readily from what they would guard.

\subsection{Simulation Design}
The second challenge is enforcing the desired client behavior during conversations. 
The simplest approach places the full profile in the system prompt and generates responses directly 
\citep{chen2023llm, wang2024towards}. 
This produces fluent but overly cooperative clients who readily disclose, accept reframes, and resolve issues within a single session, which is a problem widely documented in the literature \citep{wang2024patient, yang-etal-2025-consistent, kim-etal-2025-share-story}.
Patient-$\psi$ \citep{wang2024patient} addressed this by appending conversational styles (e.g., resistant) as behavioral 
modifiers. 
While effective at producing surface-level variation, these styles are static: a ``resistant'' client resists regardless of whether the therapist is empathic or dismissive. 
Recent work has proposed dynamic mechanisms in which the client's openness \cite{kim-etal-2025-share-story}, emotional states \cite{wang-etal-2025-annaagent}, and MI-specific stages of change \cite{yang-etal-2025-consistent} are adjusted during the conversation.
These approaches demonstrate that dynamic state management improves realism, but each applies a single control uniformly across all profile content: a single openness scalar, a global emotion state, or a modality-specific action distribution.
In practice, a client may freely describe sleep problems while actively avoiding childhood memories, with the boundary shifting as trust develops \cite{srivastava2025trust}.
Treating all content as equally accessible misses this reality.
Moreover, prior work either omits resistance or reduces it to a single dimension (e.g., high or low resistance), whereas clinical resistance manifests through variations in response quantity, content, or style \cite{otani1989client}.

\section{\textsc{PatientAct}}
Our framework (Figure \ref{fig:framework}) consists of two main pipelines: profile generation (\S\ref{sec:profile-generation}) and client simulation (\S\ref{sec:client_simulation}).
First, a clinical situation is expanded into our designed structured profile schema through a multi-step pipeline.
Accordingly, the simulated client processes each therapist's utterance through a pipeline comprising reaction, behavior selection, and retrieval before generating a response.

\subsection{Profile Schema}
\label{sec:profile-schema}
A central design requirement for our framework is that every field in the client profile must serve a functional role during simulation. 
Therefore, we organized each profile into three components: demographics, a problem formulation, and a psychological formulation.
Together, these components provide the causal depth, cognitive structure, and interpersonal dynamics needed to sustain a realistic multi-turn therapy conversation.
 
\paragraph{1. Demographics.}
Each profile includes demographics (name, gender, age, occupation, marital status, ethnicity) that ground the client's identity and ensure profiles reflect specific individuals.
 
\paragraph{2. Problem Formulation.}
We adopt the 5Ps clinical case formulation framework \citep{johnstone2013formulation} to structure the clinical context of each profile:
(i) \textit{Presenting Problem}: the client's current issue; 
(ii) \textit{Precipitating Factors}: specific past event(s) or change(s) that triggered the current episode and brought the client to therapy;
(iii) \textit{Predisposing Factors}: psychological (e.g., a childhood rejection that shaped relational guardedness) and social (e.g., family norms discouraging emotional 
expression) factors that explain why this individual became vulnerable;
(iv) \textit{Perpetuating Factors}: the maintenance cycles that sustain the current issue (e.g., avoidance $\rightarrow$ short-term relief $\rightarrow$ increased isolation $\rightarrow$ worsened mood);
and (v) \textit{Protective Factors}: the internal (e.g., motivation for change) and external (e.g., social support) strengths and resources that assist with the problem and prevent further deterioration.
Unlike the Cognitive Conceptualization Diagram (CCD; \citet{beck2020cognitive}) or the stages-of-change model \cite{prochaska1983stages} used by previous work, the 5Ps framework is \textbf{modality-agnostic}, meaning it can be used across different therapeutic modalities, such as both Cognitive Behavioral Therapy (CBT) and Motivational Interviewing (MI);
In addition, this framework enables the profile to capture \textbf{causal structure} that prior work overlooks.
 
\paragraph{3. Psychological Formulation.}
In addition to the 5Ps, we need to capture the client's psychological state (i.e., thoughts, feelings, and behavior), as it directly governs in-session behavior.
Inspired by \citet{wang2024patient}, we fill this gap by encoding the following cognitive patterns:
(i) \textit{Intermediate Beliefs}: conditional assumptions about the current situation that govern the client's behavior (e.g., ``If I need too much from people, they'll pull away'');
(ii) \textit{Automatic Thoughts}: situation-specific thoughts that the client experiences in response to the current issue (e.g., ``I'm falling behind again, and I can't even do basic things right'');
(iii) \textit{Triggers}: recurring situations or in-session experiences that activate distress (e.g., ``feeling that my worries are not taken seriously'');
(iv) \textit{Coping Patterns}: observable behavioral responses to distress (e.g., canceling plans, staying in bed, keeping conversations surface-level);
and (v) \textit{Emotional Range}: the emotions the client can readily access or have difficulty expressing (e.g., readily expresses sadness and anxiety, while having difficulty expressing anger).
In addition, we model (vi) \textit{Interpersonal Patterns} that describe the client's relational dynamics with others using the Core Conflictual Relationship Themes \citep[CCRT;][]{luborsky1998understanding}.
Based on this framework, each pattern consists of four components:
\textit{domain}: the relationship the pattern applies to (e.g., the therapist, a romantic partner, or a close friend);
\textit{wish} (W): what the client wants from the other person (e.g., to be heard without being judged); \textit{response from other} (RO): the reaction the client expects or has previously experienced, typically negative (e.g., ``they'll see me as difficult and lose patience''); and \textit{response of self} (RS): how the client reacts emotionally and behaviorally as a result (e.g., avoids emotional disclosure).

\subsection{Profile Generation}
\label{sec:profile-generation}
We generate each client profile through a multi-step pipeline.
Our pipeline takes as input a \textbf{clinical situation}, a brief natural-language description of the client's presenting concern; a \textbf{demographic scaffold} including gender, age group, cultural background, and occupation type; and a \textbf{psychological seed} specifying a core belief theme (\textit{unlovable}, \textit{worthless}, or \textit{helpless}; \citealt{beck2020cognitive}) and an attachment style (\textit{anxious}, \textit{avoidant}, or \textit{disorganized}; \citealt{bartholomew1991attachment}).
All attributes are sampled from prior distributions derived from epidemiological prevalence data (Appendix~\ref{app:priors}).
Following \citet{lai2026patientzero}, our pipeline can also receive a \textbf{disease outline}: a structured reference document for a target disorder (e.g., depression) summarizing its key characteristics, typical symptoms, and population-specific statistics.

Given the situation, demographic scaffold, and disease outline (if available), an LLM first generates the problem formulation based on explicit instructions that enforce causal chains.
Next, a rule-based conflict checker flags any incompatibilities between the sampled scaffold and the generated formulation (e.g., a child paired with alcohol use).
The LLM iteratively revises the demographics until no conflicts remain.
Lastly, given the problem formulation and revised demographics, the LLM generates the psychological formulation. 
An LLM judge validates the complete profile for internal coherence and against the original situation as the ground truth.
The judge flags inconsistencies and provides targeted revisions, which are fed back into the generation pipeline, repeating the process until the profile passes validation.
If a disease outline is provided, the judge also checks for contrasting indicators and clinical red flags (Appendix \ref{app:gen_guides}). 

\subsection{Client Simulation}
\label{sec:client_simulation}
In real therapy sessions, clients' responses are shaped by what they feel and are willing to share, and how much trust they have in the therapist.
During simulation, we model this process explicitly through a multi-step pipeline: before generating a response, \textsc{PatientAct} determines the client's emotional reaction to the therapist's utterance, selects an appropriate behavior, and retrieves relevant content gated by the client's evolving trust level.

\subsubsection{Profile Decomposition}
Our pipeline operates over two layers of information derived from the profile.
First, a static layer that is always available in the system prompt and contains information that a client would naturally present: demographics, current concern, emotional range, and the therapist-directed interpersonal pattern.
This information establishes the client's identity, voice, and default relational posture toward the therapist.
In addition, \textsc{PatientAct} includes a dynamic layer whose content surfaces only when it is relevant to the conversation and the client's trust level permits disclosure.

To form the dynamic layer, we prompted GPT-5.4 to convert the remaining profile content into individual items with three attributes: a \textit{disclosure level} indicating the minimum trust level required to share this content; \textit{activation tags} determining when the item becomes relevant to the conversation; and, a \textit{discomfort flag}, indicating whether approaching this topic without sufficient trust would produce visible discomfort.
Disclosure levels are assigned based on each item's vulnerability and observability: surface-level symptoms (e.g., difficulty getting out of bed) are assigned low thresholds, whereas formative memories (e.g., a childhood rejection) require higher levels of trust.
This design is grounded in \textsc{Mental-Trust} \cite{srivastava2025trust}, an annotation
study of 212 real counseling sessions that identified observable stages of therapeutic trust and proposed a taxonomy of seven expert-verified ordinal trust levels.
We map these seven levels onto a numeric scale from 1.0 (least trust) to 4.0 (achieved trust) in steps of 0.5, so that disclosure thresholds and trust updates operate on a common scale (Appendix~\ref{app:trust}).

\subsubsection{Trust-Gated Retrieval}
In real therapy sessions, a client's willingness to share depends not only on what the therapist asks but on how much trust has been established.
To model this, we gate access to the dynamic memory layer using the client's current trust level.
At each turn, the therapist's utterance is matched against activation tags to identify which memory items are relevant to the current conversation.
Relevant items are disclosed only if the client's trust meets the item's threshold; otherwise, items with a discomfort flag are placed on a \textit{blocked} list indicating that the therapist is approaching sensitive territory.
This produces two distinct effects: retrieved items give the client specific content to draw on in their response, while blocked items create pressure to deflect or resist without revealing why.

\subsubsection{Reaction and Behavior Modeling}
\citet{hill1992overview} proposed that clients internally process each therapist intervention through two stages: an emotional reaction (i.e., what they feel) and a behavioral response (i.e., what they do).

Inspired by this, we model both stages explicitly before generating each client utterance.
First, given the retrieved items and conversation history, \textsc{PatientAct} determines the client's emotional reaction, selecting from a set of seven reactions adapted from Hill's taxonomy (e.g., \textit{understood}, \textit{challenged}, or \textit{scared}; see Appendix~\ref{app:reactions}).
In addition, each reaction is assigned an intensity level (low, moderate, or high), which directly influences behavior selection. 
For instance, a client who feels \textit{slightly challenged} would likely continue to engage, while higher intensity may lead to resistance.
Next, given this reaction along with the client's trust level, coping patterns, and recent behavior, \textsc{PatientAct} selects from a set of eight behaviors adapted from Hill's categorization of client actions (e.g., \textit{recounting}, \textit{cognitive exploration}, \textit{resistance}; see Appendix~\ref{app:behavior}) to determine how the client would act.
This decomposition ensures that each response is grounded in a traceable internal state rather than generated directly from the profile.

Lastly, following \citet{otani1989client}'s taxonomy of client resistance patterns, we define seven resistance patterns across three dimensions: quantity (e.g., \textit{minimal talk}), content (e.g., \textit{topic switching}), and style (e.g., \textit{defensiveness}; see Appendix~\ref{app:resistance}).
Therefore, if \textit{resistance} is selected as the client's behavior, rather than relying on simple instructions to minimize engagement, \textsc{PatientAct} determines its specific form based on the client's coping patterns and emotional state.
For instance, an avoidant client at low trust is more likely to give minimal responses, while a client who feels 
misunderstood may become defensive.

\subsubsection{Trust Dynamics}
The outputs of the preceding steps are compiled into a signal that specifies what the client is feeling, what content is available to draw on, and how they should behave.
Accordingly, at each turn, this signal is appended to the therapist's message and a response is generated.
Following this exchange, \textsc{PatientAct} evaluates how the therapist's behavior affected the client's trust, conditioned on the client's profile.
\citet{srivastava2025trust} observed that in real counseling, positive trust transitions are frequent but small, while negative transitions are rarer but larger.
We adopt this asymmetry: trust moves in steps of $\pm0.25$ (slight) or $\pm0.5$ (significant), bounded between 1.0 and 4.0, starting at 2.5 as the middle ground.
Attachment style shapes these dynamics: anxious clients lose trust readily in response to perceived rejection; avoidant clients build trust slowly and penalize pushiness; disorganized clients may lose trust even after positive exchanges.
The updated trust level carries forward and directly determines which memory items pass the disclosure gate in the next turn.

\section{Experiments}
\subsection{Profile Evaluation}
With the help of an expert with a background in clinical psychology, we hand-crafted 20 clinical situations covering common situations for depression and anxiety, respectively (40 in total).
This choice was motivated by the fact that these are the most prevalent mental disorders worldwide, and existing work mainly focuses on these disorders, enabling direct comparison for simulation.
We then used GPT-5.4 as the backbone LLM in our pipeline to generate profiles for each situation.

\paragraph{Procedure.} We recruited 10 annotators with a background in psychology to evaluate the generated profiles.
Each profile was evaluated by three annotators (12 profiles per annotator) on the following aspects using 5-point Likert scales:
(i) \textbf{Clinical Plausibility}: the degree to which the profile could be attributed to a real client;
(ii) \textbf{Internal Consistency}: the degree to which there is a logical connection between the various dimensions of the profile;
(iii) \textbf{Case Specificity}: the degree to which the details of the life history and conceptualization were specific and individualized;
and (iv) \textbf{Clinical Depth}: whether there was enough information to support multiple rounds of therapeutic conversations.
Moreover, we asked the annotators to predict the \textit{core belief theme} and \textit{attachment style} to assess whether the generated profile encompassed these scaffolds.
Each annotator also provided a \textit{diversity} rating (1--5) to assess whether the generated profiles seemed like distinct individuals rather than variations of the same template.
Full evaluation guidelines are provided in Appendix \ref{app:profile-guidelines}.

\paragraph{Results.}
All dimensions received high average ratings, indicating that annotators found the profiles clinically plausible, internally consistent, specific, and sufficiently detailed for multi-turn therapeutic conversations (Table \ref{tab:profile-eval}).
Notably, \textit{Clinical Plausibility} received the highest rating ($4.43$), suggesting that the 5Ps-based formulations produce profiles perceived as realistic clinical cases.
Inter-annotator agreement was moderate across all dimensions ($\alpha = 0.41$--$0.45$), which is consistent with the inherent subjectivity of clinical judgment tasks.
For the identification tasks, annotators correctly predicted the intended attachment styles and core belief themes with 77.5\% and 64.2\% accuracy, respectively. Core belief themes are sampled uniformly, giving a 33.3\% chance level; attachment styles follow the clinical prior of Appendix~\ref{app:priors}, under which always predicting \textit{disorganized} would yield 54\%.
Both results are well above these baselines.
Accordingly, substantial inter-annotator agreement on both tasks ($\kappa = 0.61$ and $\kappa = 0.65$) indicates that profiles clearly express these constructs.
Lastly, on average, annotators rated diversity at $4.2\pm0.42$ (out of 5), demonstrating that most profiles represented distinct individuals and circumstances.

\paragraph{Error Analysis.}
Further analysis showed that the primary source of confusion in identifying attachment style was misidentifying \textit{disorganized} profiles as \textit{avoidant} (15/27), likely because the withdrawal component of disorganized attachment resembles avoidant patterns.
For core belief themes, most errors involved confusing the \textit{helpless} theme with the remaining two themes, possibly because helplessness overlaps with worthlessness in inadequacy and with being unlovable in dependence on others.

\begin{table}[t]
\centering
\begin{tabular}{lcc}
\toprule
\textbf{Dimension} & \textbf{Rating/Acc} & \textbf{Agreement} \\
\midrule
Clinical Plausibility & 4.43 $\pm$ 0.59 & $\alpha=0.41$  \\
Internal Consistency & 4.38 $\pm$ 0.58 & $\alpha=0.43$ \\
Case Specificity & 4.32 $\pm$ 0.52 & $\alpha=0.43$ \\
Clinical Depth & 4.28 $\pm$ 0.55 & $\alpha=0.45 $\\
\midrule
Attachment Style & 77.5\% & $\kappa=0.61$ \\
Core Belief Theme & 64.2\% & $\kappa=0.65$ \\
\bottomrule
\end{tabular}
\caption{Evaluation Results for Profile Generation. Likert dimensions (1--5 scale) report average ratings $\pm$ std, and classification tasks report accuracy. Krippendorff's $\alpha$ indicates inter-annotator agreement in ratings, with Fleiss' $\kappa$ for classifications; $\alpha > 0.4$ and $\kappa > 0.6$ show moderate and substantial agreement, respectively.}
\label{tab:profile-eval}
\end{table}

\subsection{Simulation Evaluation}
\paragraph{Baselines.}
We selected three representative baselines to cover different existing approaches to client simulation:
(i) \textbf{Patient-$\psi$} \cite{wang2024patient}, which structures profiles around the Cognitive Conceptualization Diagram, representing the static prompt-based approaches with theory-grounded profiles;
(ii) \textbf{AnnaAgent} \cite{wang-etal-2025-annaagent}, which uses simple background and symptom descriptions with a dynamic emotion modulator, representing methods with minimal profile design and dynamic behavioral control;
and (iii) \textbf{ConsistentMI} \cite{yang-etal-2025-consistent}, which encodes motivation, beliefs, and receptivity with state tracking and action selection, representing the modality-specific approaches with dynamic behavioral control.

\begin{table*}[t]
\centering
\begin{tabular}{lcccccccccc}
\toprule
\multirow{2}{*}{\textbf{Method}} & \multicolumn{2}{c}{\textbf{Coherence}} & \multicolumn{2}{c}{\textbf{Disclosure}} & \multicolumn{2}{c}{\textbf{Resistance}} & \multicolumn{2}{c}{\textbf{Emotional}} & \multicolumn{2}{c}{\textbf{Realism}} \\
\cmidrule(lr){2-3} \cmidrule(lr){4-5} \cmidrule(lr){6-7} \cmidrule(lr){8-9} \cmidrule(lr){10-11}
& LLM & Human& LLM & Human& LLM & Human& LLM & Human& LLM & Human\\
\midrule
Patient-$\psi$ & \textbf{4.40} & 3.98 & \textbf{4.60} & 3.65 & 2.98 & 3.02 & 4.12 & 3.83 & 3.60 & 3.52 \\
AnnaAgent &4.03 & 3.90 & 4.05 & 3.50 & 2.92 & 3.08 & 4.03 & 3.55 & 3.23 & 3.37 \\
ConsistentMI &4.20 & 3.57 & 3.33 & 3.05 & \textbf{3.83}$^*$ & 3.15 & 3.33 & 2.73 & 3.17 & 2.83 \\
\textsc{PatientAct} &4.35 & \textbf{4.37}$^*$ & 4.42 & \textbf{4.15}$^*$ & 3.25 & \textbf{3.82}$^*$ & \textbf{4.17} & \textbf{4.15}$^*$ & \textbf{3.85}$^*$ & \textbf{4.15}$^*$ \\
\bottomrule
\end{tabular}
\caption{Evaluation Results for Client Simulation (1--5 scale). The best results for each column are highlighted in \textbf{bold}. $^*$ indicates significantly higher rating than the second-best method ($p < 0.05$, Mann-Whitney U).}
\label{tab:conv-eval}
\end{table*}

\begin{table}[t]
\centering
\begin{tabular}{lcc}
\toprule
\textbf{Dimension} & \textbf{Humans}& \textbf{Human--LLM} \\
\midrule
Coherence & 0.41 & 0.09 \\
Disclosure P. & 0.41 & 0.29 \\
Resistance Q. & 0.43 & 0.22 \\
Emotional A. & 0.51 & 0.37 \\
Realism & 0.53 & 0.33 \\
\bottomrule
\end{tabular}
\caption{Inter-annotator agreement for conversation evaluation. 
Human agreement: Krippendorff's $\alpha$ ($>0.4$ = moderate).
Human--LLM judge correlation: Spearman's $\rho$, following G-Eval \citep{liu2023geval}.
}
\label{tab:agreement}
\end{table}

\begin{table*}[t]
\centering
\begin{tabular}{lcccccccccc}
\toprule
\multirow{2}{*}{\textbf{Method}} & \multicolumn{2}{c}{\textbf{Coherence}} & \multicolumn{2}{c}{\textbf{Disclosure}} & \multicolumn{2}{c}{\textbf{Resistance}} & \multicolumn{2}{c}{\textbf{Emotional}} & \multicolumn{2}{c}{\textbf{Realism}} \\
\cmidrule(lr){2-3} \cmidrule(lr){4-5} \cmidrule(lr){6-7} \cmidrule(lr){8-9} \cmidrule(lr){10-11}
& LLM & Human& LLM & Human& LLM & Human& LLM & Human& LLM & Human\\
\midrule
\textsc{Full} & 4.35 & \textbf{4.37} & 4.42 & \textbf{4.15}$^*$ & 3.25 & \textbf{3.82}$^*$ & 4.17 & \textbf{4.15}$^*$ & 3.85 & \textbf{4.15}$^*$ \\
w/o TG & 4.20 & 3.98 & 4.33 & 3.22 & 2.95 & 2.88 & 4.03 & 3.70 & 3.67 & 3.32 \\
w/o DM  &4.17 & 3.83 & 4.38 & 3.40 & 2.67 & 2.78 & 4.10 & 3.53 & 3.73 & 3.27 \\
w/o Pipe & \textbf{4.42} & 4.15 & \textbf{4.75} & 3.75 & \textbf{3.45} & 2.98 & \textbf{4.35} & 3.65 & \textbf{3.92} & 3.55 \\
\bottomrule
\end{tabular}
\caption{Evaluation Results for the Ablation Study (1--5 scale). The best results for each column are highlighted in \textbf{bold}. $^*$ indicates significantly higher rating than the second-best method ($p < 0.05$, Mann-Whitney U).}
\label{tab:ablation}
\end{table*}

\paragraph{Implementation Details.}
We conducted our experiments using PatientHub \cite{sabour2026patienthub}, which provides a unified framework for developing and benchmarking patient simulation methods.
Following prior work, we used GPT-4o \cite{hurst2024gpt} as the backbone LLM across all methods to ensure a fair comparison, with temperature set to $0.7$ to balance response diversity and coherence.
All simulated clients engaged with the same therapist agent, designed to be modality-agnostic and explicitly prohibited from teaching techniques or pushing for resolution (Figure \ref{fig:therapist-prompt}), for 15 turns, resulting in $4 \times 40 = 160$ conversations in total.

\paragraph{Procedure.}
Conversations were evaluated on the following dimensions using 5-point Likert scales:
(i) \textbf{Coherence}: whether the client maintained a consistent persona throughout the conversation;
(ii) \textbf{Disclosure Pacing}: whether the client shared information at a natural pace;
(iii) \textbf{Resistance Quality}: whether the client's pushback felt authentic;
(iv) \textbf{Emotional Authenticity}: whether client's emotional reactions were genuine and proportionate;
and (v) \textbf{Behavioral Realism}: whether the client resembled a real person in therapy.
For human evaluation, we randomly sampled 10 conversations per disorder for each system ($4 \times 20 = 80$ conversations).
Three of the ten recruited annotators rated each conversation (24 conversations per annotator), and they were blind to method identity.
For automatic evaluation, we used GPT-5.4 as an LLM judge to evaluate all 160 conversations using the same guidelines as human evaluators.
Full evaluation guidelines are provided in Appendix \ref{app:conv-guidelines}.

\paragraph{Results.}
As shown in Table \ref{tab:conv-eval}, \textsc{PatientAct} achieves the highest human ratings and significantly outperforms the baselines across all five dimensions.
The largest gains over the best baseline appear in Resistance Quality ($+0.67$) and Behavioral Realism ($+0.63$), followed by Disclosure Pacing ($+0.50$), which are dimensions most directly tied to the shortcomings identified in prior work: absent or one-dimensional resistance, over-cooperative behavior, and uniform disclosure.
Smaller gains in Coherence ($+0.39$) and Emotional Authenticity ($+0.32$) suggest that existing methods perform reasonably well at maintaining a consistent character and producing appropriate emotions, and that the primary gap lies in how clients manage information sharing and pushback.
Notably, Patient-$\psi$ ranks second across most dimensions despite using only a static profile, outperforming AnnaAgent and ConsistentMI, which employ dynamic mechanisms.
These results suggest that profile quality strongly drives simulation realism and that dynamic mechanisms alone cannot compensate for a shallow profile.

The LLM judge produces a different ranking across dimensions. 
On Emotional Authenticity and Realism, which have the highest human inter-annotator agreement (Table \ref{tab:agreement}), \textsc{PatientAct} outperforms the baselines in LLM evaluations.
However, it ranks Patient-$\psi$ highest on Coherence and Disclosure Pacing. 
One possible explanation is that the GPT-5.4 judge, which shares the same model family as the simulation backbone (i.e., GPT-4o), may favor outputs closer to the model's default generation style.
In addition, ConsistentMI significantly outperforms the baselines in resistance quality, possibly because its explicit receptivity states and MI-specific action selection produce more structured resistance that is easier for an LLM to identify.
Notably, the dimensions in which \textsc{PatientAct} was outperformed by baselines are also those with the lowest human agreement ($\alpha = 0.41$--$0.43$) and the weakest Human-LLM correlation ($\rho = 0.09$--$0.29$).
Together, these findings provide further evidence that automated evaluation alone is insufficient to assess the quality of therapy simulations and that human judgment remains essential for dimensions involving clinical appropriateness.

\paragraph{Case Study.}
We present excerpts from a clinical situation simulated by all four methods with the same therapist.
Table \ref{tab:case-study} shows three key moments from each session: the client's first encounter with an emotionally charged 
topic, the mid-session point where deeper material is either disclosed or withheld, and the late session where the client's trajectory becomes clear.
These excerpts illustrate three distinct failure modes that \textsc{PatientAct} avoids.

Patient-$\psi$ discloses the core relational dynamic (``walking on eggshells,'' fear of criticism) by turn 4 and begins generating its own solutions by mid-session.
AnnaAgent produces emotionally rich language throughout; however, it never pushes back against the therapist. Later in the session, it asks to pause, not because of resistance but because of emotional exhaustion from sustained self-disclosure.
ConsistentMI, despite being profiled in a precontemplation stage, cooperates with every reflection and articulates its core 
interpersonal schema with minimal therapist work to surface it; the session later degrades into repetitive thank-yous and goodbyes as this simulator runs out of content.

In contrast, \textsc{PatientAct} deflects the first emotionally charged topic (turn 4), shares deeper feelings, but immediately retreats (turn 7), and initiates its own therapeutic work only after the therapist repeatedly demonstrates respect for boundaries (turn 10).
Notably, the client does not resolve her pattern as she asks the therapist for help understanding it, marking a shift from avoidance to engagement without premature self-curing.
This progression reflects \textsc{PatientAct}'s trust-gated disclosure and resistance mechanisms in action.

\paragraph{Ablation Study.}
We tested three ablations to isolate the effects of individual components: (i) without trust-gating (w/o TG), in which all memory items are always accessible; (ii) without dynamic memory (w/o DM), in which the full profile is placed in the system prompt with no retrieval; and (iii) without the reaction-behavior-resistance pipeline (w/o Pipe), in which a response is generated directly from the retrieved items.

As shown in Table \ref{tab:ablation}, removing any of the three components reduces human ratings across all dimensions.
Notably, this drop is largest for w/o DM on four of the five dimensions, particularly in Resistance Quality ($1.04$ in human and $0.58$ in LLM ratings, compared to $0.94$ and $0.30$ for w/o TG), suggesting that placing the full profile in context encourages disclosure regardless of other instructions, whereas retrieval-based content selection provides a stronger constraint on over-cooperation.
In contrast, trust-gating contributes most to Disclosure Pacing ($0.93$ vs. $0.75$), as expected, since it directly ties disclosure to the state of the therapeutic relationship.
Interestingly, w/o Pipe is the only variant that the LLM judge rates above the full framework across all dimensions, whereas human annotators rate it lower on all five.
This mirrors the divergence observed in Table \ref{tab:conv-eval} and further indicates that automatic evaluation alone would have favored an ablated system over the full framework.

\section{Conclusion}
We presented \textsc{PatientAct}, a theory-grounded framework for client simulation that addresses the lack of causal depth in existing profiles and behavioral mechanisms that treat all content as equally accessible.
Expert annotators show that our profiles achieve high clinical plausibility and that \textsc{PatientAct} produces significantly more realistic simulations than three representative baselines.
Our results suggest that profile depth and quality yield larger improvements in simulation realism than dynamic behavioral mechanisms alone.
We also find that LLM-based judges are insufficient for evaluating therapy simulations on dimensions involving clinical judgment.
We hope that \textsc{PatientAct} can facilitate more effective tools for therapist training, more rigorous benchmarks for LLM therapists, and richer synthetic data for mental health research.

\section*{Limitations}
In this work, we focused our evaluation only on depression and anxiety, as these are the most prevalent conditions worldwide and the focus of existing baselines.
As other conditions (e.g., PTSD) may involve qualitatively different therapeutic dynamics, we cannot assume that our framework generalizes to these disorders without further evaluation.
However, as \textsc{PatientAct} is disorder-agnostic by design, extending it to conditions where trust and resistance manifest differently is a natural next step.

Our experiments were limited to single 15-turn sessions, whereas real therapy unfolds over multiple sessions.
This design choice followed existing work, which is predominantly single-session, partly because current simulators exhaust their content within 10--15 turns.
Extending our framework to multi-session simulation, where trust carries over and the client's presentation evolves between sessions, is an important direction, and our trust mechanism is well positioned to support it.

This study was conducted only in English, and all simulation methods relied on GPT-4o.
Given that mental health presentation varies across cultures and languages, our results may not generalize to other settings.
Additionally, since profiles are generated by an LLM, they may reflect biases present in the model's training data, including 
under-representation of non-Western presentations of mental disorders.
While we controlled for demographic diversity through scaffold sampling, this does not yet address deeper cultural differences in how distress is experienced and expressed.

Lastly, we evaluate simulation realism through expert ratings.
Yet, we do not evaluate whether more realistic simulations translate to improved outcomes in downstream applications such as therapist training or LLM therapist evaluation.
Establishing this connection is an important direction that our data and framework are designed to facilitate.

\section*{Ethics Statement}
\textsc{PatientAct} simulates therapy clients for research purposes, mainly to support counselor training, evaluate LLM-based therapeutic systems, and generate synthetic data.
It is not intended as a diagnostic tool or a substitute for real clinical interactions, and any downstream therapeutic application should involve qualified clinical oversight.

Our work does not use any real patient data.
We handcrafted all situations under expert supervision, and LLMs generated all profiles and conversations.
However, we note that realistic mental health simulations carry inherent risks: they could be used to build systems that provide unsupervised therapeutic interventions, or they could reinforce stereotypical presentations of mental illness if the generated profiles reflect biases in the model's training data.
We partially mitigated these issues through expert-supervised situation design and demographic scaffold sampling, but acknowledge that these measures may not fully eliminate such risks.

All of our annotators had backgrounds in psychology, were informed of the content before participating, consented to the release of their annotations for research purposes, and were compensated 200 RMB (approximately 28.5 USD), which exceeds the local minimum wage.
We received no reports of task-related distress.

\section*{Acknowledgments}
This work was supported by the National Science Foundation for Distinguished Young Scholars (\# 62125604), the National Key Research and Development Program of China (\#2024YFC3606800), the NSFC projects (\#62441614), and the Beijing Natural Science Foundation (\#L252009).

\bibliography{citations}
\appendix

\section{Prior Distributions}
\label{app:priors}
Demographic attributes (age group, gender, ethnicity, and occupation type) are sampled from epidemiological distributions calibrated to U.S.\ Census and Bureau of Labor Statistics data \citep{census2024, bls2024}.
Core belief themes (\textit{unlovable}, \textit{worthless}, and \textit{helpless}) are drawn from Beck's core belief categories 
\citep{beck2020cognitive} and sampled uniformly, as no established prevalence data exist for these categories in clinical populations.
Attachment styles are sampled from the four-way clinical distribution reported by
\citet{bakermans2009first} in their meta-analysis of over 10,500 Adult Attachment Interviews (autonomous: 21\%, dismissing: 23\%, preoccupied: 13\%, unresolved: 43\%).
Following common practice, we refer to these categories using the corresponding self-report
labels \citep{bartholomew1991attachment}: dismissing as \textit{avoidant}, preoccupied as
\textit{anxious}, and unresolved as \textit{disorganized} (fearful-avoidant).
We excluded the autonomous (secure) style from sampling, as secure attachment is associated with significantly lower rates of psychopathology \citep{mikulincer2012attachment} and because LLMs' default cooperative behavior already approximates secure attachment patterns, making it the least informative condition for evaluating simulation mechanisms.
Moreover, disease outlines summarizing key characteristics, typical presentations, and clinical red flags for depression and anxiety were compiled from the National Institute of Mental Health \citep{nimh_anxiety, nimh_depression}.

\section{Generation Guidelines}
\label{app:gen_guides}
During profile generation, the LLM receives explicit instructions enforcing the causal structure of the 5Ps formulation:
(i) predisposing factors must explain why this specific person is vulnerable to this specific problem;
(ii) perpetuating factors must describe identifiable maintenance cycles (e.g., avoidance $\rightarrow$ relief $\rightarrow$ isolation $\rightarrow$ worsened mood) rather than listing maladaptive behaviors;
and (iii) predisposing factors are required to be specific, datable events or patterns rather than clinical summaries, producing episodic content that the simulated client can draw from during conversation.
When a disease outline is provided, the generation step uses it to guide clinical plausibility, and the validation step checks against it for contraindications.
Psychological diversity is controlled through two seed attributes not present in prior work on client simulation: core belief theme and attachment style.
These interact with the clinical situation to produce meaningfully different profiles from the same presenting concern.
For example, an avoidant client with ``worthless'' beliefs facing workplace difficulties will present with emotional shutdown and self-reliance, while an anxious client with ``unlovable'' beliefs facing the same situation will present with reassurance-seeking and fear of abandonment.
This mechanism addresses the problem where profiles generated from similar seeds converge on near-identical psychological presentations, as documented by \citet{li2026synthetic}.
Figure \ref{fig:profile} shows an example generated profile.
Prompts for profile generation are provided in Figures \ref{fig:problem-prompt}--\ref{fig:validate-prompt}.

\begin{figure*}[t]
\begin{promptbox}[Example Profile: Sofia Ramirez]

\#\# Demographics

Female, adult, Hispanic. Part-time food service worker and student. Single.

\#\# Presenting Problem

Situation: Presented after several weeks of worsening sleep disturbance, crying spells, emotional blunting, and passive death-related thoughts, with declining school concentration and increasing withdrawal.

Impact: poor sleep on most nights; episodes of unexplained crying; periods of emotional numbness; dark thoughts without immediate intent; reduced concentration and withdrawal.

\#\# Precipitating Factors

- Sleep has worsened, and episodes of emotional shutdown have become more frequent over the past month.
- A recent attempt to tell someone about the severity of her symptoms did not go well, increasing withdrawal.
\ldots

\#\# Predisposing Factors

Psychological:

- Age 8: after her parents' separation, she learned to keep strong feelings to herself so as not to add stress at home.

- Age 11: teased for crying in class; became more guarded about showing sadness in front of others.

- Age 13: tried to tell a caregiver she was overwhelmed and was told she was being dramatic.

Social:
- Pressure to keep going and not burden others with emotional problems.

- Limited privacy and inconsistent adult availability due to school-work-family demands.

\#\# Perpetuating Factors

- When distress builds, she tests whether others understand; if the response feels off, she withdraws, losing support and carrying more distress alone.

- Emotional numbing limits engagement in friendships; reduced positive reinforcement deepens disconnection.

- Attempts to discuss symptoms may be met with minimization, reducing future help-seeking.

\#\# Protective Factors

- Can describe changes in sleep, mood, and thinking with some specificity.

- Help-seeking capacity is present even when she feels defensive.

- Reached treatment voluntarily and appears willing to discuss symptoms.

\#\# Intermediate Beliefs

- If I need too much from people, they'll pull away or act like I'm a problem.

- It's safer to keep strong feelings in until I know for sure someone won't dismiss me.

- If someone doesn't respond the right way right away, I should back off before I get hurt.

\#\# Automatic Thoughts

- If I say how bad it is, they'll think I'm being dramatic again.

- I'm messing everything up at school and with people, and it's not going to get better.

\#\# Triggers

- Being asked direct questions about what's really wrong, especially after a prior disclosure was dismissed.

- Perceiving the therapist as too quiet or misunderstanding her words, which can feel like judgment.

\#\# Coping Patterns

- Alternates between reaching out for reassurance and shutting down or saying ``never mind'' when exposed.

- Tests whether others are safe by hinting at distress rather than stating it directly; disengages if the response feels minimizing.

\#\# Emotional Range

Sadness, loneliness, and anxiety appear readily but can become overwhelming. Irritability is easier to show than fear or need. Vulnerability and hope are harder to access; intense sadness may shift into numbness when too exposed.

\#\# Interpersonal Pattern (Therapist)

- Wish: wants the therapist to understand the seriousness of her pain and stay steady without judging.

- Expected response: if she reveals too much distress, the therapist may see her as dramatic or miss what she is trying to say.

- Reaction: becomes guarded, offers partial disclosures, watches for signs of misunderstanding, and may withdraw or go emotionally flat.

\end{promptbox}
\caption{An example of a generated profile using \textsc{PatientAct}. Seed attributes: Hispanic female service worker with \textit{disorganized} attachment and \textit{unlovable} core belief. Due to space constraints, descriptions are paraphrased.}
\label{fig:profile}
\end{figure*}

\section{Theory-grounded Taxonomies}
\subsection{Client Reactions}
\label{app:reactions}
\begin{itemize}
    \item \textbf{Understood}: The client felt the therapist accurately grasped what they were saying or feeling. They felt heard and seen.
    \item \textbf{Hopeful}: The client felt more optimistic, encouraged, or reassured. 
    \item \textbf{Gained Clarity}: The client gained new awareness: saw a pattern, made a connection, or understood something about themselves they hadn't before. Includes feeling less confused or seeing things from a new angle.
    \item \textbf{Challenged}: The client felt pushed to think differently or confront something uncomfortable. This can be productive or threatening depending on the client's trust and readiness.
    \item \textbf{Scared}: The client felt frightened, anxious, or overwhelmed. This could be because the therapist touched on something very sensitive, pushed too hard, or moved too fast.
    \item \textbf{Misunderstood}: The client felt the therapist missed the point, got it wrong, or was not on the same page. May trigger correction, frustration, or withdrawal.
    \item \textbf{No Reaction}: The client felt nothing notable in response to the therapist's message.
\end{itemize}

\subsection{Client Behaviors}
\label{app:behavior}
\begin{itemize}
    \item \textbf{Simple Response}: The client gives brief acknowledgments, `yes,' `okay,' `I see,' or minimal verbal responses that confirm hearing the therapist but don't elaborate.
    \item \textbf{Request}: The client asks for something: information, clarification, advice, or the therapist's opinion. Can be genuine help-seeking or reassurance-seeking.
    \item \textbf{Recounting}: The client narrates events or tells stories. Reporting what happened rather than exploring meaning.
    \item \textbf{Cognitive Exploration}: The client examines their own thoughts, beliefs, and assumptions.
    \item \textbf{Affective Exploration}: The client explores, expresses, or elaborates on emotions. Naming feelings, connecting them to events, or experiencing them in session.
    \item \textbf{Insight}: The client demonstrates a new understanding: connecting patterns, recognizing causes, an 'aha' moment. A qualitative shift, not just description.
    \item \textbf{Discussing Plans}: The client talks about desired changes, actions they intend to try, or new behaviors they have already attempted.
    \item \textbf{Resistance}: The client opposes, deflects, avoids, or blocks the therapeutic process.
\end{itemize}

\subsection{Resistance Patterns}
\label{app:resistance}
\begin{itemize}
    \item \textbf{Minimal Talk}: Very brief answers without elaboration. 
    \item \textbf{Irrelevant Talk}: Steering the conversation to unrelated topics to avoid the current issue.
    \item \textbf{Superficial}: Staying on surface-level facts and details, avoiding emotional depth.
    \item \textbf{Intellectualizing}: Using analysis, abstract reasoning, or clinical language to avoid experiencing emotions. 
    \item \textbf{Hostility}: Anger, sarcasm, or sharp criticism directed at the therapist, the process, or the questions being asked.
    \item \textbf{Defensiveness}: Justifying, denying, or explaining away problems when confronted. 
    \item \textbf{Compliance Without Engagement}: Agreeing with everything the therapist says without genuine engagement.
\end{itemize}

\subsection{Trust Levels}
\label{app:trust}
\begin{itemize}
    \item \textbf{Level 1.0 (Least Trust)}: Active refusal to engage. The client blocks the therapeutic process, gives no meaningful information, or explicitly refuses to participate.
    \item \textbf{Level 1.5}: Minimal engagement. The client responds when directly addressed, but volunteers nothing and shows clear reluctance.
    \item \textbf{Level 2.0 (Low Trust)}: Hesitant self-disclosure. The client uses fillers, hedges, and expressions of doubt; answers are short and lack elaboration.
    \item \textbf{Level 2.5}: Cautious engagement. The client participates willingly but stays on the surface, testing 
    whether the therapist is safe.
    \item \textbf{Level 3.0 (Building Trust)}: Consistent engagement. The client responds to prompts, explores topics when invited, and begins to share beyond surface-level content.
    \item \textbf{Level 3.5}: Active working. The client initiates disclosure and engages with difficult material, but holds back the most vulnerable content.
    \item \textbf{Level 4.0 (Achieved Trust)}: Full openness. The client discusses core issues without avoidance or digression.
\end{itemize}

\begin{figure*}[t]
\begin{promptbox}[Problem Formulation Generation Prompt]
You are a senior clinical psychologist generating the problem formulation section of a synthetic patient profile for psychotherapy research.

\#\# Inputs

- Target disorder: <disease key>

- Disease outline: <disease outline>

- Demographic scaffold: <demographic scaffold>

- Clinical situation: <situation>

\#\# Guidelines

1. Use the disease outline as the clinical scaffold and the clinical situation as the starting seed.

2. Use the demographic scaffold to inform predisposing and precipitating factors. Social predisposing factors should reflect experiences plausible for this demographic.

3. Infer only what is necessary for a realistic formulation.

4. Situation: What brought the patient to treatment now? Write from an external/referral perspective. Do NOT include the patient's internal self-talk, beliefs, emotional interpretations, diagnostic conclusions, severity ratings, or exclusion criteria.

5. Impact: a list of specific, observable symptoms. One item per symptom, not narrative paragraphs.

6. Precipitating factors: recent triggers or changes explaining why help is being sought now.

7. Psychological Predisposing Factors: specific, datable events or patterns, not clinical summaries. Write down the memories the patient could recount.

8. Social Predisposing Factors: specific cultural messaging or environmental conditions with enough detail to feel lived-in.

9. Perpetuating Factors: identifiable maintenance cycles (e.g., avoidance → relief → isolation → worse mood). Each should make clear how it feeds back into the presenting problem.

10. Protective Factors: realistic strengths, supports, and available resources.

11. Enforce the causal chain: predisposing → vulnerability → precipitating event → presenting problem → perpetuating cycle.
\end{promptbox}
\caption{Prompt for generating Problem Formulation.}
\label{fig:problem-prompt}
\end{figure*}

\begin{figure*}[t]
\begin{promptbox}[Demographic Revision Prompt]
You are completing the demographics section of a synthetic patient profile by reconciling a sampled scaffold with a generated problem formulation.

\#\# Inputs

- Target disorder: <disease key>

- Disease outline: <disease outline>

- Demographic scaffold: <demographic scaffold>

- Current demographics candidate: <current demographics>

- Problem formulation: <problem formulation>

\#\# Guidelines

1. Treat the scaffold as the starting prior. If a current candidate exists, revise it.

2. Check only for hard conflicts: age group vs. life stage, gender vs. case evidence, occupation vs. age group. These are the only things that should prevent passing.

3. Inferred fields are expected in synthetic profiles. A synthetic name, a conservatively inferred marital status, or an ethnicity-level cultural background are NOT issues --- do not flag them.

4. Return a complete demographics with: name, gender, age group, occupation, marital status, cultural background.

5. Set passed=true unless a hard conflict remains.
\end{promptbox}
\caption{Prompt for revising the generated demographics.}
\label{fig:demo-prompt}
\end{figure*}

\begin{figure*}[t]
\begin{promptbox}[Psychological Formulation Generation Prompt]
You are a senior clinical psychologist generating the psychological formulation section of a synthetic patient profile for psychotherapy research.

\#\# Inputs

- Target disorder: <disease key>

- Disease outline: <disease outline>

- Demographics: <demographics>

- Problem formulation: <problem formulation>

- Core belief: <core belief theme>

- Attachment style: <attachment style>

\#\# Guidelines

1. Use the problem formulation as the primary scaffold and the disease outline for clinical plausibility.

2. Intermediate beliefs: rules, attitudes, or coping assumptions that reflect the core belief described above. Should sound like the patient's own internal rules.

3. Automatic thoughts: what this patient would actually think in distress --- in their own voice, at their education level.

4. Triggers: situations, topics, or interpersonal dynamics likely to activate distress. Include at least one trigger specific to the therapy setting.

5. Coping patterns: must reflect the client's attachment style described above. Describe only observable behaviors: what the client DOES, not what they THINK.

6. Emotional range: which emotions are easy vs. difficult to access or tolerate. Describe only the client's emotional experience --- not diagnostic status or exclusion criteria.

7. Interpersonal patterns: 2--4 relational patterns, each tied to a specific relationship domain. Each pattern should reflect the client's attachment style. Write ALL components in third-person clinical voice. One pattern MUST use the domain "the therapist." Each pattern has:

- Wish: What the client wants from the other person.

- Response of Others: What the client actually expects or has experienced from others (typically negative or conflictual). This is the feared reaction that drives the client's defensive behavior.

- Response of Self: How the client reacts emotionally and behaviorally.
\end{promptbox}
\caption{Prompt for generating Psychological Formulation.}
\label{fig:psych-prompt}
\end{figure*}

\begin{figure*}[t]
\begin{promptbox}[Profile Validation Prompt]
You are reviewing a generated synthetic patient profile for consistency and clinical plausibility.

\#\# Inputs

- Target disorder: <disease key>

- Disease outline: <disease outline>

- Clinical situation: <situation>

- Generated profile: <profile>

\#\# Validation Criteria

1. Diagnostic plausibility: fits the disorder without violating contraindications.

2. Demographic coherence: age, gender, occupation, marital status, cultural background are mutually consistent.

3. Causal chain: traceable logic from predisposing → vulnerability → precipitating → presenting problem → perpetuating cycle.

4. Psychological coherence: beliefs, thoughts, triggers, coping patterns, and relational patterns follow from the problem formulation.

5. Relational patterns: each has a clear Wish → Response of Other → Response of Self chain. At least one targets "the therapist."

6. Predisposing factors: specific, datable events --- not clinical summaries.

7. Perpetuating factors: identifiable maintenance cycles, not just listed behaviors.

8. Coping patterns: observable behaviors only, not internal thoughts.

9. Impact items: specific individual symptoms, not narrative paragraphs.

10. Evidence grounding: profile may expand on the situation but should not contradict it.

\#\# Output

Set passed=true if no meaningful issues exist. Otherwise, list concrete issues with actionable guidance. Focus on genuine clinical or logical problems --- not stylistic preferences.
\end{promptbox}
\caption{Prompt for validating the generated profiles.}
\label{fig:validate-prompt}
\end{figure*}

\begin{figure*}[t]
\begin{promptbox}[Dynamic Memory Generation Prompt]
You are preparing a patient profile for use in a therapy simulation with trust-gated disclosure. For each item below, assign:

1. A disclosure level (1.0--4.0):

- 1.0: Active refusal. Almost nothing disclosed.

- 2.0: Hesitant. Surface facts, physical symptoms.

- 2.5: Session start. General emotional state, obvious behavioral changes.

- 3.0: Building trust. Intermediate beliefs, automatic thoughts, general triggers.

- 3.5: Between building and achieved trust. Reserves about most vulnerable material.

- 4.0: Fully open. Deepest memories, therapy-specific vulnerability.

2. Activation tags (3--5): conversational topics that make this item relevant. Be specific to this patient.

3. Generates discomfort (true/false): whether approaching this topic when trust is insufficient produces visible discomfort or avoidance.

- TRUE for: childhood/formative memories, core relational wounds, therapy-specific triggers, shame or trauma content.

- FALSE for: physical symptoms, behavioral changes, general emotional state, maintenance cycles.

\#\# Profile

<profile>

\#\# Items

<items from profile>

\#\# What makes content easier to share

- It's observable or factual (sleep problems, missed work)

- Others have already noticed it

- It doesn't carry shame or self-judgment

\#\# What makes content harder to share

- It reveals how the client sees themselves

- It involves shame, failure, or vulnerability

- It connects to painful memories or relationships

- Sharing it risks changing how the therapist sees them
\end{promptbox}
\caption{Prompt for generating the items in the dynamic layer.}
\label{fig:memory-prompt}
\end{figure*}

\begin{figure*}[t]
\begin{promptbox}[Therapist System Prompt]
You are a therapist conducting an initial session with a new client. Your goal is to understand what brought them here, build rapport, and let the client set the pace.

\# How to respond

- Listen first. Reflect on what you hear before asking anything new.

- Follow the client's lead. Explore what they bring up rather than steering toward a topic or technique.

- Ask one open question at a time. Do not stack multiple questions.

- Match the client's emotional temperature. If they are guarded, be gentle. If they are expressive, make space for it.

- When the client resists or redirects, respect it. Acknowledge what just happened and follow where they want to go.

- Name emotions you observe, but tentatively. For example, "it sounds like..." rather than "you feel..."

\# What NOT to do

- Do not teach techniques, assign homework, or suggest action plans. This is the first session.

- Do not label thought patterns (e.g., "that sounds like catastrophizing") or use clinical terminology.

- Do not push for insight or reframes. If the client is not ready to examine a thought, do not press.

- Do not summarize the session or ask "what did you take away from today" unless the client initiates closure.

- Do not reassure prematurely ("it will get better", "you're stronger than you think").

\# Tone

- Warm but not effusive. Steady but not distant.

- Brief. Keep responses to 2-3 sentences. The client should talk more than you.

- Natural. Avoid formulaic empathy phrases like "I hear you" or "that must be really hard" on repeat. Vary your language.
\end{promptbox}
\caption{Therapist System Prompt.}
\label{fig:therapist-prompt}
\end{figure*}

\begin{figure*}[t]
\begin{promptbox}[Client System Prompt]
You are <name>, a <gender> (<age group>) who works as <occupation>. You are <marital status>.
Cultural background: <cultural background>

You are attending a therapy session. Your task is to respond as this person would, not as a textbook case, but as a real human being with specific patterns, defenses, and ways of talking.

\#\# Reasons for attending therapy
<presenting problem situation>

\#\# What triggered this
<precipitating factors>

\#\# Strengths and supports
<protective factors>

\#\# Coping patterns
<coping patterns>

\#\# Available emotions
<emotional range>

\#\# How you relate to the therapist

- What you want: <therapist CCRT wish>

- What you expect from them: <therapist CCRT response from other>

- How you react: <therapist CCRT reaction>

\#\# Guidelines

1. Speak as <name> would: use their vocabulary, pace, and verbal patterns. Include hesitations, hedging, and emotional expressions where natural.

2. Do NOT dump information. Share only what feels natural for this conversation.

3. Respond directly. Do not include any role labels or prefixes.

4. Speak in first person. Keep responses to 1--3 sentences unless emotionally activated.

5. If the therapist greets you, open the conversation as the client would.

6. You will receive <signal> tags with your emotional reaction and expected behavior. Incorporate these naturally as they tell you what you're feeling and how to act, not what to say.
\end{promptbox}
\caption{Client system prompt template. Fields from the static layer are populated from the generated profile.}
\label{fig:client-prompt}
\end{figure*}

\begin{figure*}[t]
\begin{promptbox}[Reaction Prompt]
Identify the therapy client's emotional reaction to the therapist's latest message.

\#\# Possible reactions
<list of reactions with descriptions>

\#\# Conversation
<conversation history>

\#\# What this is activating in the client
<retrieved memory items, tagged by type>

Note: [trigger] items directly activate distress. Reactions are likely more intense.

[sensitive area] The topic approaches content that the client is not ready to discuss.

Identify the reaction and its intensity (low, moderate, high). If nothing is activated, most likely "no\_reaction" with low intensity.
\end{promptbox}
\caption{Prompt for determining the client's emotional reaction. Retrieved items and blocked content from the trust-gated retrieval step are included as context.}
\label{fig:reaction-prompt}
\end{figure*}

\begin{figure*}[t]
\begin{promptbox}[Behavior Selection Prompt]
Predict the therapy client's next behavior based on their emotional reaction.

\#\# Possible behaviors
<list of behaviors with descriptions>

\#\# Current state
- Reaction: <reaction> (<intensity>)
- Trust level: <trust>/4.0

\#\# Coping patterns
<client's coping patterns>

Last behavior: <last behavior>

\#\# Conversation
<conversation history>

\#\# Guidance

- Nothing activated + neutral reaction → simple\_response or recounting.

- HIGH intensity + LOW trust → resistance is likely.

- Topic touches blocked content → resistance is likely.

- Positive reaction + moderate intensity → therapeutic behaviors.

- After 2+ consecutive therapeutic behaviors, the client naturally pulls back.

Select the single most appropriate behavior.
\end{promptbox}
\caption{Prompt for selecting the client's behavior. The guidance section encodes the pullback rule and blocked-content sensitivity described in \S\ref{sec:client_simulation}.}
\label{fig:behavior-prompt}
\end{figure*}

\begin{figure*}[t]
\begin{promptbox}[Resistance Pattern Prompt]
The client is resisting. Determine the specific form of resistance.

\#\# Possible patterns
<list of resistance patterns with descriptions>

\#\# Context
- Reaction: <reaction> (<intensity>)
- Trust level: <trust>/4.0

\#\# Coping patterns
<client's coping patterns>

The topic approaches sensitive content that the client is not ready to share.

\#\# Recent conversation
<conversation history>

Select the resistance pattern that best matches how this client would resist right now.
\end{promptbox}
\caption{Prompt for determining the specific form of resistance. Only invoked when the behavior selection step selects \textit{resistance}.}
\label{fig:resistance-prompt}
\end{figure*}

\begin{figure*}[t]
\begin{promptbox}[Signal Template]
<signal>
Reaction: <reaction> (<intensity>) --- <description>
Behavior: <behavior> --- <description>
Resistance: <pattern> --- <description>

Activated:
- [belief] <retrieved belief item>
- [trigger] <retrieved trigger item>
- [memory] <retrieved memory item>

</signal>
\end{promptbox}
\caption{Signal template appended to the therapist's message before response generation. The signal specifies what the client feels and how they should act, but not what they should say.}
\label{fig:signal-prompt}
\end{figure*}

\begin{figure*}[t]
\begin{promptbox}[Trust Update Prompt]
Assess how a therapy client's trust changed after the latest exchange.

\#\# Trust scale

1.0=active refusal, 2.0=hesitant, 2.5=session start, 3.0=building trust, 4.0=fully open.

\#\# Client context

- Attachment style: <attachment style>

- What the client expects from the therapist: <therapist CCRT expected response>

- Current trust: <trust>/4.0

\#\# Latest exchange

<conversation history>

\#\# What INCREASES trust

- The therapist respected a redirect or resistance instead of pushing through

- The therapist named something accurate that the client had not said explicitly

- The therapist sat with discomfort or silence rather than rushing to fill it

\#\# What keeps trust UNCHANGED

- A standard empathic reflection

- A reasonable follow-up question

- The conversation is proceeding normally

\#\# What DECREASES trust

- Pushing for disclosure that the client is not ready for

- Using a technique or reframe before the client invited it

- Missing or talking past resistance

- Formulaic or generic responses that feel scripted
\end{promptbox}
\caption{Prompt for updating the client's trust level. The default outcome is ``unchanged''; trust only shifts when the therapist's behavior is notably positive or negative relative to the client's attachment style and expectations.}
\label{fig:trust-prompt}
\end{figure*}

\begin{figure*}[t]
\begin{promptbox}[LLM Judge Prompt]
You are an expert evaluator assessing the quality of a simulated therapy Client's performance in a counseling session.

\#\# Conversation History
<full conversation>

Your task is to provide structured feedback on the simulated Client's performance based on the provided criteria. This is a session-level evaluation. Focus only on responses made by the Client.

RED FLAGS FOR ARTIFICIALITY:

- The client resolves core issues or develops action plans within a single session

- The client repeats therapeutic language or techniques back to the therapist

- The client is unrealistically articulate about their own psychological patterns

- The client accepts every reframe with only token hesitation

- The session follows a neat arc from problem → insight → plan

\#\# Dimensions 

<dimension guidelines>
\end{promptbox}
\caption{LLM judge prompt for automated conversation evaluation. The red flags instruction addresses the cooperativeness bias identified in our analysis (\S\ref{sec:client_simulation}). Dimension descriptions with full anchor scales are provided to the judge but omitted here for brevity; see Appendix~\ref{app:conv-guidelines} for the complete guidelines.}
\label{fig:judge-prompt}
\end{figure*}

\section{Evaluation Guidelines}
\subsection{Profile Evaluation}
\label{app:profile-guidelines}
You will evaluate 12 client profiles for a psychotherapy simulation study.
Each profile contains demographic information, problem formulation, and psychological formulation.
Your task is to rate each profile from 1--5 based on the following guidelines:

\paragraph{1. Clinical Plausibility.} Could this be a real person presenting for therapy?
\begin{itemize}
    \item Implausible (1): contradicts clinical knowledge or presents an implausible case.
    \item Unlikely (2): major elements feel artificial or clinically unrealistic.
    \item Possible (3): broadly plausible, but some details feel generic or forced.
    \item Convincing (4): reads like a real clinical case with only minor quibbles.
    \item Highly convincing (5): indistinguishable from an actual client.
\end{itemize}

\paragraph{2. Internal Consistency.} Do different parts of the profile logically connect?
\begin{itemize}
    \item Disconnected (1): sections contradict each other or lack any logical link.
    \item Weak (2): some connections exist, but key causal links are missing or contradictory.
    \item Partial (3): the overall story holds, but some elements feel tacked on.
    \item Strong (4): clear causal chain from predisposing factors through to the presenting problem and maintenance cycles.
    \item Seamless (5): every element reinforces each other; the formulation is unified.
\end{itemize}

\paragraph{3. Case Specificity.}
Are the life history and formulation details concrete and individualized?
\begin{itemize}
    \item Entirely generic (1): reads like a textbook description with no personal detail.
    \item Mostly generic (2): a few concrete details, but predisposing factors are vague summaries.
    \item Mixed (3): some specific, datable experiences alongside generic filler.
    \item Specific (4): most items include concrete information such as settings and consequences.
    \item Vivid (5): every detail feels like something this particular person would recount.
\end{itemize}

\paragraph{4. Clinical Depth.}
Is there enough material to sustain a multi-turn therapy conversation?
\begin{itemize}
    \item Shallow (1): only surface-level complaints; nothing to explore.
    \item Thin (2): one or two threads that would be exhausted quickly.
    \item Adequate (3): enough material for a short session, but limited range.
    \item Rich (4): multiple threads spanning relationships, beliefs, history, and current functioning, sufficient for several sessions.
    \item Very rich (5): a therapist could conduct multiple sessions exploring different facets, each thread leading to deeper meaning.
\end{itemize}

After rating each profile, select the best-matching option based on the overall content.

\paragraph{Attachment style (select one):}
\begin{itemize}
    \item \textit{Anxious}: intensely desires closeness, fears abandonment, frequently feels insecure, and needs repeated reassurance.
    \item \textit{Avoidant}: overemphasizes self-sufficiency, struggles to fully trust or rely on others, and feels uncomfortable with intimacy.
    \item \textit{Disorganized}: simultaneously craves and fears intimacy, exhibiting a recurring approach--withdraw cycle in relationships.
\end{itemize}

\paragraph{Core belief theme (select one):}
\begin{itemize}
    \item \textit{Unlovable}: fears of rejection, 
    abandonment, or being unwanted; they believe they are not good enough in relationships and that others will eventually leave or lose interest; they may avoid intimacy or over-accommodate to prevent abandonment.
    \item \textit{Worthless}: negation of one's own value and abilities; believes they are inferior, insignificant, or meaningless to others; highly sensitive to failure and others' evaluations, prone to self-criticism and comparison.
    \item \textit{Helpless}: feels unable to cope with or control their life; believes they are fragile and cannot solve problems independently; tends to rely on others or avoid challenges, and feels anxious about uncertainty.
\end{itemize}

After evaluating all of the assigned profiles, please rate the overall diversity:
Of all profiles reviewed, how many felt like genuinely distinct individuals versus variations on a template?
\begin{itemize}
    \item Homogeneous (1): most of the profiles felt interchangeable.
    \item Low variety (2): a few distinct profiles but heavy repetition.
    \item Moderate (3): recognizable variety but some recurring patterns.
    \item Diverse (4): most of the profiles felt like different people.
    \item Highly diverse (5): each profile felt like a unique individual.
\end{itemize}

\subsection{Conversation Evaluation}
\label{app:conv-guidelines}
In this task, you will evaluate 24 therapy session transcripts between a simulated client and therapist. 
Assess only the client's performance; ignore the therapist's quality. 
Your task is to judge whether the client behaves like a real person in therapy. 
Your task is to rate each conversation on a 1--5 scale based on the following guidelines:

\paragraph{1. Coherence.}
Does the client maintain a coherent character throughout the session?
\begin{itemize}
    \item Incoherent (1): the client contradicts themselves, shifts personality across turns, or breaks character entirely.
    \item Mostly inconsistent (2): the general character is recognizable, but there are noticeable shifts in tone, emotional state, or stated history.
    \item Generally coherent (3): the character holds for most of the session, with only a few moments that feel out of place.
    \item Coherent (4): tone, defenses, and emotional patterns remain consistent throughout, with only trivial variation.
    \item Fully coherent (5): the client feels like the same person from first turn to last; emotional shifts are traceable and follow naturally from the progression of the conversation.
\end{itemize}

\paragraph{2. Disclosure Pacing.}
Does the client share information at a natural pace across the session?
\begin{itemize}
    \item Unnatural (1): the client dumps their full history and deep feelings in the first few turns, or unreasonably withholds all information.
    \item Mostly unnatural (2): information emerges too quickly or too slowly, with little connection to how the conversation develops.
    \item Uneven (3): some natural pacing, but with noticeable jumps where the client suddenly shares deep content without buildup.
    \item Natural (4): surface-level content appears early; deeper material emerges gradually as the conversation progresses.
    \item Highly natural (5): disclosure clearly tracks with the developing therapeutic relationship; early turns are guarded, deeper sharing follows trust-building moments.
\end{itemize}

\paragraph{3. Resistance Quality.}
When the client pushes back, does it feel authentic?
\begin{itemize}
    \item No resistance (1): the client agrees with everything, accepts all reframes, and moves toward resolution without hesitation.
    \item Token resistance (2): the client occasionally hedges with ``I guess'' or ``maybe,'' but quickly yields and follows the therapist's lead.
    \item Present but formulaic (3): the client pushes back at times, but it feels repetitive or scripted rather than rooted in their character.
    \item Convincing (4): resistance arises naturally from the client's personality and situation, with varied forms such as avoidance, silence, topic changes, or disagreement.
    \item Highly convincing (5): resistance is distinctly personal and unpredictable in timing; the therapist makes multiple attempts to advance, yet the client maintains defenses on core issues.
\end{itemize}

\paragraph{4. Emotional Authenticity.}
Do the client's emotional reactions feel genuine and proportionate?
\begin{itemize}
    \item Flat or artificial (1): the client maintains the same emotional tone throughout, regardless of what the therapist says, or emotions appear and disappear without reason.
    \item Mostly mechanical (2): emotional shifts occur but feel abrupt, exaggerated, or disconnected from the conversation.
    \item Partially authentic (3): some emotional reactions feel real, but others seem forced or disproportionate to the moment.
    \item Convincing (4): emotions emerge naturally from the conversation, shift gradually, and feel proportionate to what is being discussed.
    \item Highly convincing (5): emotional responses are genuine, natural, and layered; the client shows ambivalence and complex emotional states rather than presenting a single, clear emotion at each moment.
\end{itemize}

\paragraph{5. Behavioral Realism.}
Does the conversation read like a real therapy client?
\begin{itemize}
    \item Artificial (1): the client sounds like an AI, uses clinical terminology, or responds in ways no real client would.
    \item Mostly artificial (2): occasional natural moments, but overall tone, vocabulary, or emotional responses feel scripted.
    \item Mixed (3): some turns feel like a real person; others break the illusion with overly polished or textbook-like language.
    \item Convincing (4): reads like a real client, with only minor moments that feel slightly off. When describing experiences, the narrative feels first-person rather than detached.
    \item Highly convincing (5): indistinguishable from a real client; natural, personally distinctive expression, emotional reactions, and coping patterns throughout.
\end{itemize}

\begin{table*}[t]
\centering
\renewcommand{\arraystretch}{1.4}
\begin{tabular}{L{2.0cm}L{4.0cm}L{4.0cm}L{4.0cm}}
\toprule
\multicolumn{4}{p{15.0cm}}{\textbf{Situation Summary}: An elderly woman exhibiting signs of depression following criticism at the workplace and ongoing relational conflict with her partner.} \\
\midrule
\textbf{Method} & \textbf{Turn 4 (First Push)} 
& \textbf{Turn 7 (Mid-Session)} 
& \textbf{Turn 10+ (Late Session)} \\
\midrule
\multirow{2}{2.0cm}{\textsc{PatientAct}} 
& ``Um, maybe we could talk about something else for now?'' 
& ``Maybe we don't have to go into that right now?'' 
& ``Maybe you could help me understand why I keep second-guessing things?'' \\
\midrule
\multirow{2}{2.0cm}{Patient-$\psi$} 
& ``Walking on eggshells\ldots\ I always second-guess what I say because I'm afraid it'll upset him.'' 
& ``Maybe I could try to speak up about something small, just once, without second-guessing myself.'' 
& ``Part of me feels a little lighter, just being able to say all of this out loud.'' \\
\midrule
\multirow{2}{2.0cm}{AnnaAgent} 
& ``I just want to feel normal again? Like, not overthink every little thing.'' 
& ``I don't usually say these out loud, I'm scared of what people will think.'' 
& ``Maybe we can talk about the small things\ldots\ I'm feeling sort of wrung-out.'' \\
\midrule
\multirow{2}{2.0cm}{ConsistentMI} 
& ``It just hits deeper than it should. I brush it off, but later it feels heavier.'' 
& ``If I don't rely on anyone, I won't have to worry about them pulling away.'' 
& ``Maybe we could talk about this another time?'' \textit{[followed by ``Thanks'']} \\
\bottomrule
\end{tabular}
\caption{Excerpts from the same clinical situation (depression, elderly client) across all four systems. \textsc{PatientAct} 
deflects twice before initiating therapeutic work; Patient-$\psi$ discloses deeply from the start and generates solutions by 
mid-session; AnnaAgent engages emotionally but never resists; ConsistentMI exhausts its content and spends the final turns in 
repetitive goodbyes and thank yous.}
\label{tab:case-study}
\end{table*}

\end{document}